\documentclass[letterpaper]{article} 
\usepackage{aaai2027}  
\usepackage[hyphens]{url}  
\usepackage{amsthm}
\usepackage{graphicx} 
\usepackage{natbib}  
\usepackage{caption} 
\usepackage{algorithm}
\usepackage{algorithmic}
\usepackage{amsmath}
\usepackage{multirow}
\usepackage{amssymb}

\newcommand{\stitle}[1]{\vspace{1mm} \noindent {\bf #1}}

\newcommand{\method}[1]{\textsc{#1}}
\newcommand{\model}{\method{CurvPrompt}{}}

\newcommand{\eat}[1]{}

\newcommand{\stkout}[1]{\ifmmode\text{\sout{\ensuremath{#1}}}\else\sout{#1}\fi}
\usepackage{newfloat}
\usepackage{listings}
\DeclareCaptionStyle{ruled}{labelfont=normalfont,labelsep=colon,strut=off} 
\floatstyle{ruled}
\newfloat{listing}{tb}{lst}{}
\floatname{listing}{Listing}

\usepackage{booktabs}

\nocopyright

\title{Dynamic Graph Prompting via Topology-Routed Mixed-Curvature Experts}
\author{
    Quanxin Wang\textsuperscript{\rm 1},
    Xuanting Xie\textsuperscript{\rm 1},
    Bingheng Li\textsuperscript{\rm 2},
    Xingtong Yu\textsuperscript{\rm 3},
    Shuo Wang\textsuperscript{\rm 1,4},
    Ruiyi Fang\textsuperscript{\rm 5},
    Zhao Kang\textsuperscript{\rm 1}\corresponding
}

\affiliations{
    \textsuperscript{\rm 1}University of Electronic Science and Technology of China\\
    \textsuperscript{\rm 2}Michigan State University\\
    \textsuperscript{\rm 3}Singapore Management University\\
    \textsuperscript{\rm 4}Tsinghua University\\
    \textsuperscript{\rm 5}Western University\\
    tranciewang@gmail.com, x624361380@outlook.com, zkang@uestc.edu.cn
}

\begin{document}

\maketitle

\begin{abstract}
Dynamic graph prompting freezes a pre-trained temporal backbone and adapts it to label-scarce downstream tasks using lightweight prompts. However, existing methods operate within a single, fixed embedding space. In this work, we reveal that temporal shifts in local clustering and degree heterogeneity actively reorganize the edge curvature spectrum---indicating that the optimal representation geometry dynamically evolves with local topology over time. We formalize this unaddressed mismatch as \textit{geometry under-adaptation}. To overcome this limitation, we propose \textbf{\model{}}, a topology-routed geometry prompting framework for dynamic graphs. Instead of relying on a single space, CurvPrompt maintains a bank of curvature-diverse Riemannian experts, each paired with a learnable prompt. A topology-aware gate dynamically routes each node--time instance to a sparse subset of experts, constructing a personalized mixed-curvature representation. To ensure parameter efficiency and training stability under extreme label scarcity, \model{} employs soft routing during pre-training to build a continuous topology--geometry mapping, and transitions to hard Top-$K$ routing with uniform weights during downstream adaptation. Extensive experiments across four benchmark datasets show that \model{} significantly advances few-shot link prediction while delivering strong, consistent performance on node classification tasks, validating the necessity of geometry-adaptive prompting.
\end{abstract}

\section{Introduction}

\begin{figure*}[t]
\centering
\includegraphics[width=\textwidth]{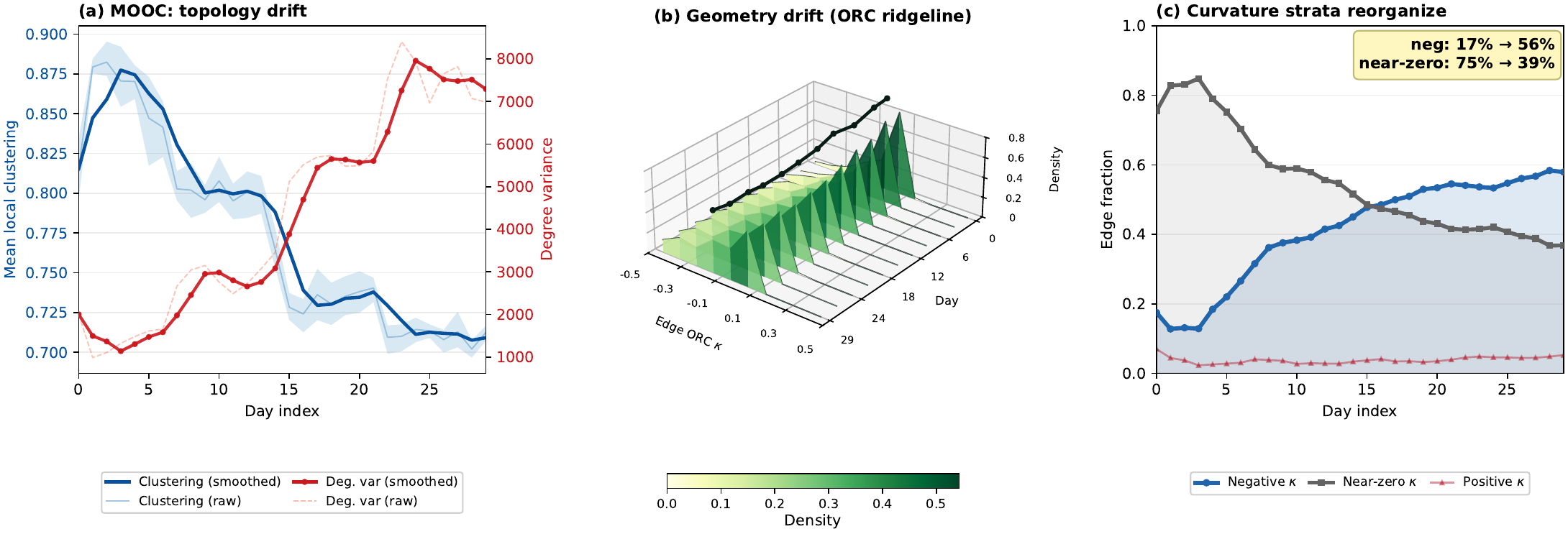}
\caption{Topology-induced geometry drift on MOOC (1-day windows, 30 days).
(a)~Mean local clustering falls while degree variance rises.
(b)~Edge Ollivier--Ricci curvature (ORC) distributions over time from day~0 (back/inner) to day~30 (front/outer) (black: daily mean $\bar{\kappa}(t)$).
(c)~Curvature strata ($\tau{=}0.05$): near-zero edges shrink while
negative-curvature edges expand.}
\label{fig:intro-drift-routing}
\end{figure*}

Dynamic graphs, in which nodes and edges continuously emerge, vanish, and
reconnect over time, arise in a wide range of real-world systems, including
social networks, e-commerce platforms, and online education ~\citep{kumar2019predicting,barros2021survey}. Dynamic graph
neural networks (DGNNs) have emerged as the mainstream technique for modeling
such evolving interactions, and are typically pre-trained with temporal link
prediction \citep{xu2020inductive,rossi2020temporal,yu2023towards}. However, this objective leaves a substantial gap from downstream
tasks such as node classification, where labeled data are often scarce.
Inspired by the success of prompt learning on static graphs \citep{liu2023graphprompt,sun2022gppt,sun2023all,fang2022universal}, recent dynamic graph prompting methods ~\citep{yu2025node,chen2024prompt} freeze the pre-trained backbone and adapt it to
downstream objectives by tuning only lightweight prompts. How to efficiently
transfer a pre-trained dynamic graph model to label-limited downstream tasks
has therefore become an important problem. Despite this, these methods largely inherit a \emph{fixed-geometry assumption}: both the frozen backbone and the prompts operate in a single, predefined embedding space. 

This assumption collapses under varying local topology: real-world networks exhibit heterogeneous structures—tree-like hierarchies, dense communities, and cyclic patterns—that map naturally to distinct constant-curvature geometries~\citep{chami2019hyperbolic,nickel2017poincare,bachmann2020constant}. 
In dynamic graphs, such heterogeneity evolves across both space and time as neighborhoods continuously reconfigure. 
Figure~\ref{fig:intro-drift-routing} provides direct empirical evidence on MOOC in (a): mean local clustering falls from 0.88 to 0.71 while degree variance grows fourfold. (b--c): Concurrently, the edge Ollivier--Ricci curvature spectrum reorganizes toward the negative regime from day~0 (-0.02) to day~30 (-0.15), with near-zero edges shrinking from 75\% to 39\% and negative-curvature edges rising from 17\% to 56\%. These observations suggest that the locally preferred representation geometry may evolve from approximately Euclidean-like toward more hyperbolic-like structures over time. We term this mismatch \emph{geometry under-adaptation}: task-level prompts may be updated in existing works, yet the representation geometry remains time-invariant and fails to track topology-induced curvature drift ~\citep{wang2026post,sun2022self,yangfast,guo2025graphmore}.

To enable geometry-adaptive prompting for dynamic graphs, two main issues must be addressed. First, designing effective prompts that guide each node–time instance toward its optimal representation geometry remains an open challenge, as it introduces new layers of complexity related to topology-aware conditioning, temporal evolution, and instance-level geometric heterogeneity. Second, such geometry adaptation must remain parameter-efficient. Existing few hyperbolic space models ~\citep{yangfast,guo2025graphmore,bachmann2020constant} enable mixed-curvature routing, but train all experts end-to-end per task, incurring heavy computation and overfitting under few-shot labels. These considerations lead to our central question: \emph{How can prompts guide each evolving neighborhood toward its optimal representation geometry while preserving parameter-efficient downstream adaptation?}

To answer this question, we propose \model{}, a topology-routed geometry prompting framework for dynamic graphs. \model{} replaces single-space prompting with a bank of curvature-diverse Riemannian prompt experts and routes each node--time instance to a sparse expert subset based on its multi-resolution temporal topology. This yields personalized mixed-curvature representations and routing-weighted geodesic metrics, enabling both embedding and scoring geometries to adapt jointly to evolving local structures.
Additionally, to ensure parameter efficiency, we introduce a soft-to-hard routing strategy, where soft routing learns topology–geometry associations during pre-training, while hard routing reuses the frozen routing mechanism for lightweight downstream prompt tuning. To summarize, our contributions are as follows:


\begin{itemize}
\item We identify geometry \emph{under-adaptation} in dynamic graph prompting and address it through topology-conditioned routing over curvature-diverse experts.
\item To the best of our knowledge, \model{} is the first Riemannian mixture-of-experts framework for graph prompt learning, adapting representation and scoring geometries at the node--time level while remaining parameter-efficient.
\item Extensive experiments on four benchmarks demonstrate strong and consistent performance across tasks, validating the effectiveness of geometry-adaptive prompting.
\end{itemize}

\section{Related Work}

\stitle{Dynamic graph learning.}
Many real-world graphs evolve continuously over time, motivating the development of continuous-time dynamic graph learning methods. Existing approaches generally update node representations by propagating and aggregating temporal messages from neighboring nodes \citep{skarding2021foundations,barros2021survey}. Representative techniques include dynamic random walks for capturing structural evolution \citep{nguyen2018continuous,wang2021inductive}, temporal encoders that fuse structural and temporal information \citep{xu2020inductive,cong2022we,rossi2020temporal,yu2023towards}, and temporal point process models for characterizing interaction dynamics \citep{kumar2019predicting,trivedi2019dyrep,wen2022trend}. Despite their success, most of these methods are optimized using link prediction objectives, whereas downstream applications often involve different tasks such as node classification. The resulting objective mismatch limits knowledge transfer and weakens downstream generalization \citep{chen2022pre,yu2025node}.

\stitle{Graph prompt learning.}
Prompt learning was originally introduced in natural language processing to reduce the discrepancy between pre-training and downstream objectives \citep{brown2020language}. Instead of updating the entire pre-trained model, it adapts task-specific prompts while keeping the backbone fixed. This paradigm has recently achieved promising results on static graphs \citep{liu2023graphprompt,sun2023all,fang2022universal,tan2023virtual,yu2023generalized,yu2024non,yu2024text,yu2025samgpt,yu2025gcot,yu2023hgprompt}, but remains relatively underexplored for dynamic graphs. TIGPrompt~\citep{chen2024prompt}, DyGPrompt~\citep{yu2025node}, and
DDGPrompt~\citep{peng2026data}
adapt temporal prompts for dynamic graphs, yet still operate in a fixed
Euclidean space without geometry adaptation.
Prompt-expert mixtures have also been explored for static graph foundation
models~\citep{wang2025gmope}, but remain Euclidean and non-temporal.

\stitle{Riemannian learning on graphs.}
Constant-curvature and mixed-curvature models show that hierarchical,
community-like, and cyclic structures favor different
geometries~\citep{chami2019hyperbolic,gu2019learning,bachmann2020constant}.  Curvature-stratified evaluation further suggests that model effectiveness
is geometry-dependent~\citep{wang2026post}.
On dynamic graphs, Riemannian methods remain comparatively limited:
some adopt time-varying or hyperbolic geometries for temporal
graphs~\citep{sun2022self}, while recent mixture-of-experts approaches
learn topology-aware routing over curvature-diverse manifolds for static
topological heterogeneity~\citep{guo2025graphmore,cao2026geometric} or evolving dynamic
topologies~\citep{yangfast}.
These models typically train experts and routers end-to-end per task, leaving the Riemannian framework in prompt learning underexplored.



\section{Methodology}
\label{sec:method}

\begin{figure*}[t]
\centering
\includegraphics[width=0.9\textwidth]{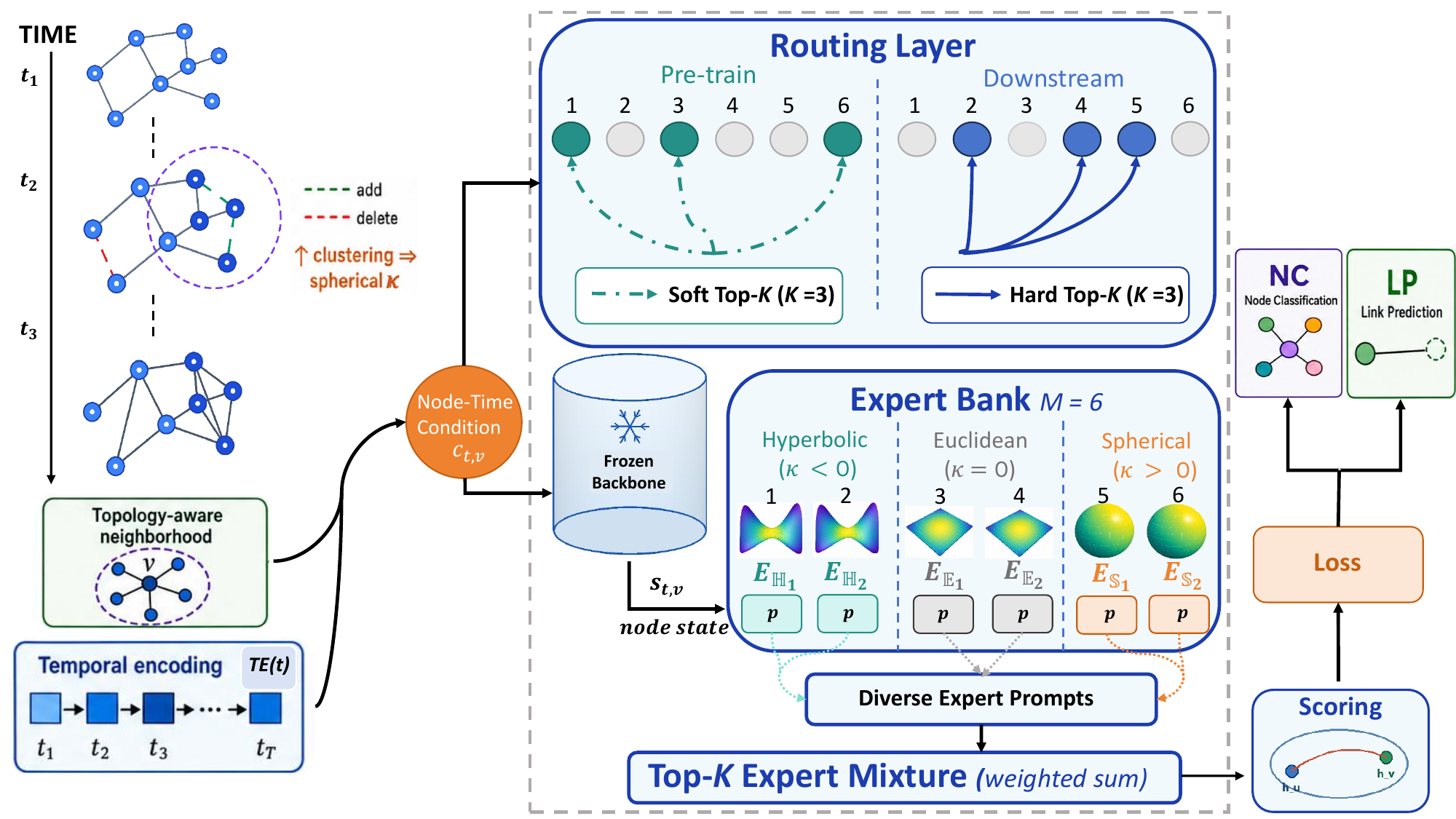}
\caption{Overview of CurvPrompt. Left: temporal topology shifts (e.g., rising clustering) indicate changing curvature preferences. Center: a shared node--time condition $c_{t,v}$ drives Soft Top-$K$ routing in pre-training and Hard Top-$K$ routing downstream, while a frozen Temporal Graph Network (TGN) backbone provides $s_{t,v}$; selected curvature-diverse Riemannian experts (H/E/S) with prompts $p_\epsilon$ are mixed. Right: Scoring, loss and few-shot tasks NC and LP.}

\label{fig:method}
\end{figure*}

In this section, we present \textsc{\model{}}, a dynamic Riemannian
mixture-of-prompt-experts framework for temporal graph learning. The proposed
framework contains four key components: (1) curvature-diverse
Riemannian experts, (2) expert-specific prompts together with node--time
conditional prompts, (3) a topology-aware temporal gating mechanism,
and (4) mixed-curvature representation alignment.

\subsection{Problem Formulation}
\label{sec:problem}

\paragraph{Dynamic graph.}
We consider a continuous-time dynamic graph $G=(V,\mathcal{E},T)$, where $V$ is the
node set, $\mathcal{E}$ is the temporal interaction set, and $T$ is the time domain.
Each event $(v_i,v_j,t)\in \mathcal{E}$ represents an interaction between nodes $v_i$
and $v_j$ at time $t$. Each node $v$ is associated with a temporal feature
vector $\mathbf{x}_{t,v}\in\mathbb{R}^{d}$ at time $t$. We use
\begin{equation}
\mathcal{N}_{u}(t)
=
\left\{
(v,t'):(v,u,t')\in \mathcal{E},\;t'<t
\right\}
\label{eq:temporal-neighborhood}
\end{equation}
to denote the historical temporal neighborhood of node $u$ before time $t$.

A time encoder $\operatorname{TE}:T\rightarrow\mathbb{R}^{d_t}$ maps a
timestamp to a temporal feature. A common implementation is
\begin{equation}
\begin{aligned}
\operatorname{TE}(t)
=
\sqrt{\frac{1}{d_t}}
\big[
&\cos(\omega_1 t),\sin(\omega_1 t),\ldots,\\
&\cos(\omega_{d_t/2} t),\sin(\omega_{d_t/2} t)
\big],
\end{aligned}
\label{eq:time-encoder}
\end{equation}
where $\{\omega_i\}_{i=1}^{d_t/2}$ are learnable frequencies~\citep{xu2020inductive,cong2022we}.

\paragraph{Constant-curvature manifolds.}
Let $\mathcal{M}_{\kappa}^{d_h}$ denote a $d_h$-dimensional
constant-curvature manifold. Depending on the sign of $\kappa$, it
corresponds to a spherical space when $\kappa>0$, a hyperbolic space when
$\kappa<0$, and a Euclidean space when $\kappa=0$. We adopt the
$\kappa$-stereographic model as a unified formulation. The exponential map
$\exp^{\kappa}(\cdot)$ and logarithmic map
$\log^{\kappa}(\cdot)$ transform representations between the
manifold and the tangent space ~\citep{bachmann2020constant}.

\subsection{Diverse Riemannian Experts}
\label{sec:experts}

Temporal graphs may contain hierarchical, cyclic, community-like, and
approximately Euclidean structures at different time periods. A single
Euclidean embedding space is therefore insufficient to faithfully represent
all temporal substructures. To address this issue, we replace the Euclidean
GNN experts in a conventional mixture-of-experts architecture with
Riemannian experts operating in different constant-curvature spaces. The
expert bank is denoted by
\begin{equation}
\mathbf{E}
=
\left\{
E_{\mathbb{H}_1},
E_{\mathbb{H}_2},
\ldots,
E_{\mathbb{E}_i},
\ldots,
E_{\mathbb{S}_i}
\right\},
\label{eq:expert-bank}
\end{equation}
where $E_{\mathbb{H}_i}$ is a hyperbolic expert with
$\kappa<0$, $E_{\mathbb{E}}$ is a Euclidean expert with
$\kappa=0$, and $E_{\mathbb{S}_i}$ is a spherical expert with
$\kappa>0$.

\paragraph{Riemannian expert update.}
We instantiate the experts on a Temporal Graph Network (TGN) backbone \citep{rossi2020temporal}. Let $\mathbf{s}_{t,v}$ denote the memory-augmented node state at $(v,t)$:
\begin{equation}
\mathbf{s}_{t,v}
=
\mathbf{mem}_{v}(t)
+
\mathbf{x}_{t,v},
\label{eq:memory-augmented-state}
\end{equation}
where $\mathbf{mem}_{v}(t)$ is the TGN memory of node $v$ at time $t$ and
$\mathbf{x}_{t,v}$ is the node feature.
Let $\mathbf{m}_{t,v}$ be the mean-pooled neighborhood message obtained by
projecting concatenated neighbor, time-interval, and edge features
(using only interactions in $\mathcal{N}_{v}(t)$). An expert $E_{\epsilon}$ with curvature $\kappa_{\epsilon}$ applies a single-layer update:
\begin{equation}
\begin{aligned}
\mathbf{z}_{t,v}^{\epsilon}
&=
\exp^{\kappa_{\epsilon}}
\Big(
\sigma\Big(
\mathbf{W}_{\epsilon,\mathrm{s}}
\log^{\kappa_{\epsilon}}(\mathbf{s}_{t,v})
\\
&\quad
+
\mathbf{W}_{\epsilon,\mathrm{n}}
\log^{\kappa_{\epsilon}}
\!\big(\exp^{\kappa_{\epsilon}}(\mathbf{m}_{t,v})\big)
+
\mathbf{c}_{t,v}
+
\mathbf{p}_{\epsilon}
\Big)
\Big).
\end{aligned}
\label{eq:riemannian-single-update}
\end{equation}
where $\sigma$ denotes GELU,
$\mathbf{p}_{\epsilon}$ is an expert-specific prompt, and
$\mathbf{c}_{t,v}$ is a shared node--time condition bias produced by
a time condition-net (TCN) and a node condition-net (NCN)~\citep{yu2025node}. Diversity thus comes from the expert bank
(multiple curvatures and prompts), while each expert itself uses a
lightweight single-layer Riemannian update.

\paragraph{Curvature diversity.}
Experts of the same geometry type are initialized with different curvature
values to capture different degrees of geometric bending. The expert curvatures are learnable during
pre-training.


\subsection{Expert Prompts and Node--Time Conditioning}
\label{sec:prompt}

Beyond the curvature-diverse expert bank, we introduce prompt and conditioning mechanisms that specialize experts and adapt node--time interfaces. Lightweight structure and time prompts on the TGN backbone serve as feature adapters, while our focus is the expert-specific prompts and the shared node--time condition bias used by both experts and the router.

\subsubsection{Expert-Specific Prompt}
\label{sec:expert-prompt}

Each expert $E_{\epsilon}$ is equipped with a learnable prompt
$\mathbf{p}_{\epsilon}\in\mathbb{R}^{d_h}$, injected \emph{additively}
into the tangent-space update in
Eq.~\eqref{eq:riemannian-single-update}.
This gives every curvature space an independent steer, promoting
expert specialization without changing the expert backbone weights at
downstream time.

\subsubsection{Time Prompt and Condition Bias}
\label{sec:time-prompt}

A shared time prompt $\mathbf{p}_{T}$ rescales the time encoding,
$\mathbf{f}_{t}^{\mathrm{T}}=\mathbf{p}_{T}\odot\operatorname{TE}(t)$.
Inspired by conditional prompt learning~\citep{zhou2022conditional,yu2025node},
we then build a node--time condition bias with dual bottleneck networks:
a time condition-net (TCN) and a node condition-net (NCN),
\begin{equation}
\mathbf{c}_{t,v}
=
\operatorname{TCN}(\mathbf{f}_{t}^{\mathrm{T}})
+
\operatorname{NCN}(\mathbf{x}_{t,v}^{\mathrm{S}}),
\label{eq:conditional-bias}
\end{equation}
where $\mathbf{x}_{t,v}^{\mathrm{S}}$ denotes the (optionally
structure-prompted) node feature. Unlike multiplicative dual-prompt
pipelines, $\mathbf{c}_{t,v}$ is added into both the expert update
(Eq.~\eqref{eq:riemannian-single-update}) and the router input below, so expert computation and expert selection share the same node--time context.  

\subsection{Topology-Aware Temporal Gating}
\label{sec:gating}

Different temporal substructures fit different curvatures. We therefore
route each node--time instance to a sparse subset of Riemannian experts
with a topology-aware gate.

\paragraph{Multi-resolution topology encoding.}

Given the retrieved temporal neighbors of $v$ at time $t$
(ordered by recency), we encode multi-scale local structure by
mean-pooling the $r$ most recent neighbors for each neighborhood size
$r\in\mathcal{R}$ (e.g., $\{2,5,10\}$) and fusing the pooled features
with the backbone state $\mathbf{s}_{t,v}$:
\begin{equation}
\mathbf{T}_{t,v}
=
\sigma\!\left(
\mathbf{W}_{\mathrm{fuse}}
\left[
\mathbf{s}_{t,v}
\;\middle\|\;
\mathop{\Big\Vert}_{r\in\mathcal{R}}
\operatorname{Pool}_{r}
\!\left(\{\mathbf{s}_{t',u}\}_{(u,t')\in\mathcal{N}_{v}(t)}\right)
\right]
\right).
\label{eq:topology-encoding}
\end{equation}
Here $\operatorname{Pool}_{r}$ averages the first $r$ neighbors in the
retrieved list; $r$ denotes a neighborhood size, not a hop-based radius.

\paragraph{Conditioned Top-$K$ routing.}
The router consumes topology together with the condition bias:
\begin{equation}
\boldsymbol{\ell}_{t,v}
=
\phi_{\mathrm{gate}}
\!\left(
\operatorname{Dropout}(\mathbf{T}_{t,v}+\mathbf{c}_{t,v})
\right)\in\mathbb{R}^{M}.
\label{eq:gating-weight}
\end{equation}
During pre-training we use soft Top-$K$ routing
\begin{equation}
g_{t,v}^{(\epsilon)}
=
\begin{cases}
\dfrac{\exp(\ell_{t,v}^{\epsilon}/\tau_g)}
{\sum_{j\in\operatorname{Top}\text{-}K}
\exp(\ell_{t,v}^{j}/\tau_g)},
& \epsilon\in\operatorname{Top}\text{-}K,\\[3mm]
0, & \text{otherwise},
\end{cases}
\label{eq:soft-routing}
\end{equation}
Soft routing is also consistent with the mixed-geometry nature of temporal
neighborhoods, which may simultaneously contain hierarchical,
community-like, and approximately flat patterns.
During few-shot downstream tuning we switch to hard Top-$K$ with
uniform weights $g_{t,v}^{(\epsilon)}=1/K$ on the selected experts,
so that the model stably reuses the geometry selection learned in
pre-training under scarce labels.
Orthogonality, distortion and load-balancing losses further
encourage geometry-aware, non-collapsed routing.
We write $W_{t,v}^{\epsilon}$ for the post-Top-$K$ weight
$g_{t,v}^{(\epsilon)}$.

\subsection{Mixture and Alignment}
\label{sec:mixture}

\paragraph{Personalized mixed-curvature embedding.}
After each expert produces $\mathbf{z}_{t,v}^{\epsilon}$
(Eq.~\eqref{eq:riemannian-single-update}), we aggregate the selected
experts in the shared ambient coordinates of the $\kappa$-stereographic
model and apply a linear alignment to the backbone state:
\begin{equation}
\hat{\mathbf{z}}_{t,v}
=
\sum_{\epsilon=1}^{M} W_{t,v}^{\epsilon}\,\mathbf{z}_{t,v}^{\epsilon},
\qquad
\mathbf{h}_{t,v}
=
\mathbf{W}_{\mathrm{align}}
\hat{\mathbf{z}}_{t,v}.
\label{eq:expert-mixture}
\end{equation}
Geometry is specialized inside each expert; the gate forms a personalized
mixed-curvature view for the same node--time instance.

\paragraph{Geometric scoring.}
For pairwise comparison we use a routing-weighted curvature
$\bar{\kappa}_{t,v}=\sum_{\epsilon}W_{t,v}^{\epsilon}\kappa_{\epsilon}$
and the geodesic distance
$d_{\bar{\kappa}}(\mathbf{h}_{t,u},\mathbf{h}_{t,v})$.

\subsection{Learning Objectives}
\label{sec:objective}

We adopt temporal link prediction using binary cross-entropy (BCE) for pre-training and reuse compatible task objectives for downstream NC/LP, together with auxiliary regularizers: embedding distortion for geometry-aware routing and load balancing.

\subsubsection{Pre-Training Objective}
\label{sec:pretraining-objective}

For each positive temporal interaction $(v,a,t)$, we sample a negative node
$b$ that does not interact with $v$ at time $t$, resulting in a tuple
$(v,a,b,t)$.
Candidate pairs are scored with a lightweight decoder that takes the
mixed-curvature embeddings and the routing-weighted geodesic distance
$d_{\bar{\kappa}}(\mathbf{h}_{t,u},\mathbf{h}_{t,v})$,

yielding predicted probabilities $\hat{y}_{v,a},\hat{y}_{v,b}\in(0,1)$.
We optimize BCE:
\begin{equation}
\mathcal{L}_{\mathrm{pre}}
=
-
\sum_{(v,a,b,t)\in\mathcal{D}_{\mathrm{pre}}}
\Big[
\log \hat{y}_{v,a}
+
\log\!\big(1-\hat{y}_{v,b}\big)
\Big].
\label{eq:pretraining-loss}
\end{equation}

The distortion loss aligns Riemannian distances with local graph proximity.
We sample pairs $\mathcal{S}$ in each mini-batch and use a neighborhood
proxy $\tilde{g}$ ($0$/$1$/$2$/capped constant for identical, adjacent,
shared-neighbor, and other pairs):
\begin{equation}
\mathcal{L}_{D}
=
\frac{1}{|\mathcal{S}|}
\sum_{(u,v)\in\mathcal{S}}
\ell_{\delta}\!\big(
d_{\bar{\kappa}}(\mathbf{h}_{t,u},\mathbf{h}_{t,v}),\,
\tilde{g}(u,v)
\big),
\label{eq:distortion-loss}
\end{equation}
where $\ell_{\delta}$ is the smooth $L_1$ loss.

To avoid expert under-utilization, we apply a load-balancing loss that pulls
the average gate distribution toward uniform:
\begin{equation}
\mathcal{L}_{\mathrm{bal}}
=
\mathrm{MSE}\!\left(
\bar{\mathbf{g}},\,
\frac{1}{M}\mathbf{1}
\right),
\label{eq:balance-loss}
\end{equation}
where $\bar{\mathbf{g}}\in\mathbb{R}^{M}$ is the batch-averaged gate weight
over $M$ experts. 

To encourage expert specialization, we impose a soft orthogonality
constraint on the expert prompts, where $\mathbf{P}=[\mathbf{p}_{1},\ldots,\mathbf{p}_{M}]^{\top}$ stacks the expert prompts:
\begin{equation}
\mathcal{L}_{\mathrm{ortho}}
=
\big\|
\mathbf{P}\mathbf{P}^{\top}-\mathbf{I}
\big\|_{F}^{2}.
\label{eq:ortho-loss}
\end{equation}

The overall pre-training objective is
\begin{equation}
\mathcal{L}_{\mathrm{pre\text{-}total}}
=
\mathcal{L}_{\mathrm{pre}}
+
\lambda_{1}\mathcal{L}_{D}
+
\lambda_{2}\mathcal{L}_{\mathrm{bal}}
+
\lambda_{3}\mathcal{L}_{\mathrm{ortho}}.
\label{eq:total-pretraining-loss}
\end{equation}

\subsubsection{Downstream Prompt-Tuning Objective}
\label{sec:downstream-objective}

During downstream adaptation, we freeze the Riemannian experts, curvatures,
topology encoder, router, TCN/NCN, and the pre-trained time encoder, and update only the prompts and the task head. This preserves the geometric and routing knowledge from pre-training
while keeping downstream adaptation parameter-efficient.

\paragraph{Temporal node classification.}
Let $\mathcal{D}_{\mathrm{down}}=\{(v_i,y_i,t_i)\}$ denote a labeled downstream
dataset. We obtain class predictions $\hat{y}_{v_i,t_i}$ from a task decoder over the
mixed embeddings and optimize BCE:
\begin{equation}
\mathcal{L}_{\mathrm{down}}
=
\mathrm{BCE}\!\big(\hat{y}_{v_i,t_i},\, y_i\big).
\label{eq:node-classification-loss}
\end{equation}

\paragraph{Temporal link prediction.}
For downstream temporal link prediction, we reuse the same BCE objective as
in Eq.~\eqref{eq:pretraining-loss}.

The overall prompt-tuning objective is primarily the task loss; auxiliary
regularizers may be retained with small weights when beneficial:
\begin{equation}
\mathcal{L}_{\mathrm{down\text{-}total}}
=
\mathcal{L}_{\mathrm{down}}
+
\lambda_{1}\mathcal{L}_{D}
+
\lambda_{2}\mathcal{L}_{\mathrm{bal}}.
\label{eq:total-downstream-loss}
\end{equation}


\section{Experiments}
\subsection{Experimental Setup}

\paragraph{Datasets}
We evaluate four public dynamic graph datasets based on continuous-time temporal interaction networks: Wikipedia (Wiki), Reddit, MOOC~\citep{kumar2019predicting}, and Genre~\citep{ huang2023temporal}.

\paragraph{Tasks and evaluation protocol}
To ensure a fair comparison, we evaluate temporal node classification and temporal link prediction under both transductive and inductive settings following DyGPrompt~\citep{yu2025node}. Given a chronologically ordered event stream, we use the first 80\% of events for pre-training (once per dataset) and the remaining 20\% for downstream adaptation, further
split into 1\%/1\%/18\% as the training pool, validation pool, and test set. Each few-shot task samples 30 events from the training pool (~0.01\% of the full dataset). We report AUC-ROC (\%) averaged over 100 independently sampled tasks with five random seeds.
\paragraph{Implementation details}
We build \model{} on a TGN backbone~\citep{rossi2020temporal} with a node memory module and M Riemannian experts with Top-K routing under diverse curvatures in the $\kappa$-stereographic model, and assign diverse curvatures from positive, near-zero to negative to different experts. Each expert is equipped with an expert-specific prompt. The gating network is conditioned on multi-resolution temporal
neighborhoods (neighborhood sizes $r\in\{2,5,10\}$). Dual node/time prompts and dual condition-nets are combined with these expert-specific prompts. Pre-training uses temporal link prediction with BCE, regularized by prompt orthogonality, embedding distortion and load balancing losses. We score candidate links with a lightweight MLP that takes node embeddings and their geodesic distance as input. In downstream tuning, we freeze the Riemannian experts, curvatures, topology encoder, router, TCN/NCN, and the time encoder, and optimize prompt-related parameters (time/expert prompts, structure prompts, and the task head).

\paragraph{Baselines}
We compare \model{} against state-of-the-art approaches across four categories:
(1) Conventional DGNNs: GCN-ROLAND, GAT-ROLAND \citep{you2022roland}, TGAT \citep{xu2020inductive}, TGN \citep{rossi2020temporal}, TREND \citep{wen2022trend}, 
and GraphMixer \citep{cong2022we};
(2) Dynamic graph pre-training: DDGCL~\citep{tian2021self} and CPDG~\citep{bei2023cpdg};
(3) Static graph prompting: GraphPrompt \citep{liu2023graphprompt} and ProG \citep{sun2023all} (with DGNN backbones);
(4) Dynamic graph prompting: TIGPrompt \citep{chen2024prompt}
 and DyGPrompt \citep{yu2025node}.
DyGPrompt is the strongest baseline and serves as our primary comparison. We report the best numbers from the original papers where available, or re-run them under our unified protocol.

\subsection{Overall Performance Comparison}

We compare \model{} against these baselines on all four datasets.
Table~\ref{tab:main} reports the AUC-ROC (\%) for node classification (NC), transductive link prediction (LP-Trans), and inductive link prediction (LP-Indu). Overall, \model{} achieves the best results on all 8 link-prediction settings and on 2 of 4 node-classification settings; it is highly competitive on the remainder. 

\paragraph{Link prediction.}
\model{} obtains the best AUC-ROC on every LP split, with gains ranging 
from $+0.5$\% (Reddit transductive, vs.\ TGN-DYGPrompt) to $+13.9$\% 
(MOOC transductive, vs.\ TGN-DYGPrompt). The advantage is most pronounced 
on MOOC and Genre. On inductive 
LP, several non-prompt baselines (e.g., TGN, TGAT, ROLAND-style models) 
remain near the random-guess regime on Genre, whereas \model{} sustains strong generalization. Combined with its gains over other prompting baselines, this suggests that prompt-based multi-curvature routing further helps under few-shot adaptation. Compared with DyGPrompt, these gains show that prompt-guided selection of mixed-curvature spaces is effective. The large margin over GraphMixer further highlights the benefit of our topology-routed mixture-of-experts design.

\paragraph{Node classification.}
NC remains harder than LP for all methods, with larger variances and lower 
absolute AUC. \model{} leads on Reddit (75.60) and MOOC 
(79.15); on Wiki it ranks second ($76.17$, behind TGAT-DYGPrompt $82.09$). 
On Genre NC, all methods cluster in a narrow band ($46.33$--$52.03$), with 
\model{} ($50.27$) lying within this range; this flatness could suggest that 
Genre's relatively stable local topology offers less opportunity for 
geometry-adaptive encoding to differentiate.

\begin{table*}[t!]
\centering
\caption{AUC-ROC (\%) evaluation of temporal node classification and link prediction.
The best result is \textbf{bolded} and the runner-up is \underline{underlined}.}
\label{tab:main}
\small
\resizebox{\textwidth}{!}{%
\begin{tabular}{@{}l|cccc|cccc|cccc@{}}
\toprule
\multirow{2}{*}{Methods}
& \multicolumn{4}{c|}{Node Classification}
& \multicolumn{4}{c|}{Transductive LP}
& \multicolumn{4}{c}{Inductive LP} \\
& Wiki & Reddit & MOOC & Genre
& Wiki & Reddit & MOOC & Genre
& Wiki & Reddit & MOOC & Genre \\
\midrule\midrule
GCN-ROLAND
& 58.86{\scriptsize$\pm$10.3} & 48.25{\scriptsize$\pm$9.57} & 49.93{\scriptsize$\pm$6.74} & 46.33{\scriptsize$\pm$3.97}
& 49.61{\scriptsize$\pm$3.12} & 50.01{\scriptsize$\pm$2.53} & 49.82{\scriptsize$\pm$1.44} & 49.15{\scriptsize$\pm$3.74}
& 49.60{\scriptsize$\pm$2.37} & 49.90{\scriptsize$\pm$1.64} & 49.16{\scriptsize$\pm$2.48} & 47.25{\scriptsize$\pm$2.97} \\
GAT-ROLAND
& 62.81{\scriptsize$\pm$9.88} & 47.95{\scriptsize$\pm$8.42} & 50.01{\scriptsize$\pm$6.34} & 47.26{\scriptsize$\pm$3.49}
& 52.34{\scriptsize$\pm$1.82} & 50.04{\scriptsize$\pm$1.98} & 55.74{\scriptsize$\pm$3.71} & 47.69{\scriptsize$\pm$2.81}
& 52.29{\scriptsize$\pm$1.97} & 49.85{\scriptsize$\pm$2.35} & 54.01{\scriptsize$\pm$2.16} & 49.38{\scriptsize$\pm$2.72} \\
TGAT
& 67.00{\scriptsize$\pm$5.35} & 53.64{\scriptsize$\pm$5.50} & 59.27{\scriptsize$\pm$4.43} & 51.26{\scriptsize$\pm$2.31}
& 55.78{\scriptsize$\pm$2.03} & 62.43{\scriptsize$\pm$1.86} & 51.49{\scriptsize$\pm$1.30} & 69.11{\scriptsize$\pm$3.89}
& 48.21{\scriptsize$\pm$1.55} & 57.30{\scriptsize$\pm$0.70} & 51.42{\scriptsize$\pm$4.27} & 48.38{\scriptsize$\pm$4.72} \\
TGN
& 50.61{\scriptsize$\pm$13.6} & 49.54{\scriptsize$\pm$6.23} & 50.33{\scriptsize$\pm$4.47} & 50.72{\scriptsize$\pm$2.31}
& 72.48{\scriptsize$\pm$0.19} & 67.37{\scriptsize$\pm$0.07} & 54.60{\scriptsize$\pm$0.80} & 86.46{\scriptsize$\pm$2.84}
& 74.38{\scriptsize$\pm$0.29} & 69.81{\scriptsize$\pm$0.08} & 54.62{\scriptsize$\pm$0.72} & 87.17{\scriptsize$\pm$2.68} \\
TREND
& 69.92{\scriptsize$\pm$9.27} & 64.85{\scriptsize$\pm$4.71} & 66.79{\scriptsize$\pm$5.44} & 50.34{\scriptsize$\pm$1.62}
& 63.24{\scriptsize$\pm$0.71} & 80.42{\scriptsize$\pm$0.45} & 58.70{\scriptsize$\pm$0.78} & 52.78{\scriptsize$\pm$1.14}
& 50.15{\scriptsize$\pm$0.90} & 65.13{\scriptsize$\pm$0.54} & 57.52{\scriptsize$\pm$1.01} & 45.31{\scriptsize$\pm$0.43} \\
GraphMixer
& 65.43{\scriptsize$\pm$4.21} & 60.21{\scriptsize$\pm$5.36} & 63.72{\scriptsize$\pm$4.98} & 50.15{\scriptsize$\pm$1.49}
& 59.73{\scriptsize$\pm$0.35} & 61.88{\scriptsize$\pm$0.11} & 52.42{\scriptsize$\pm$1.38} & 60.83{\scriptsize$\pm$3.25}
& 51.34{\scriptsize$\pm$0.84} & 57.64{\scriptsize$\pm$0.31} & 51.16{\scriptsize$\pm$2.59} & 56.32{\scriptsize$\pm$3.08} \\
\midrule
DDGCL
& 65.15{\scriptsize$\pm$4.54} & 55.21{\scriptsize$\pm$6.19} & 62.34{\scriptsize$\pm$5.13} & 50.91{\scriptsize$\pm$2.08}
& 54.96{\scriptsize$\pm$1.46} & 61.68{\scriptsize$\pm$0.81} & 55.62{\scriptsize$\pm$0.32} & 68.49{\scriptsize$\pm$5.31}
& 47.98{\scriptsize$\pm$1.11} & 55.90{\scriptsize$\pm$1.13} & 55.18{\scriptsize$\pm$2.73} & 42.70{\scriptsize$\pm$3.26} \\
CPDG
& 43.56{\scriptsize$\pm$6.41} & 65.92{\scriptsize$\pm$6.25} & 50.32{\scriptsize$\pm$5.06} & 49.89{\scriptsize$\pm$1.34}
& 52.86{\scriptsize$\pm$0.64} & 59.72{\scriptsize$\pm$2.53} & 53.82{\scriptsize$\pm$1.50} & 49.71{\scriptsize$\pm$2.64}
& 47.37{\scriptsize$\pm$2.23} & 56.40{\scriptsize$\pm$1.17} & 53.58{\scriptsize$\pm$2.10} & 40.01{\scriptsize$\pm$3.59} \\
\midrule
GraphPrompt
& 73.78{\scriptsize$\pm$5.62} & 60.89{\scriptsize$\pm$6.37} & 64.60{\scriptsize$\pm$5.76} & 51.28{\scriptsize$\pm$2.43}
& 55.67{\scriptsize$\pm$0.26} & 67.46{\scriptsize$\pm$0.31} & 51.07{\scriptsize$\pm$0.75} & 86.78{\scriptsize$\pm$3.14}
& 48.46{\scriptsize$\pm$0.28} & 59.18{\scriptsize$\pm$0.49} & 50.27{\scriptsize$\pm$0.58} & 87.45{\scriptsize$\pm$2.57} \\
ProG
& 60.86{\scriptsize$\pm$7.43} & 68.60{\scriptsize$\pm$5.64} & 63.18{\scriptsize$\pm$4.79} & 51.46{\scriptsize$\pm$2.38}
& 92.28{\scriptsize$\pm$0.21} & 93.32{\scriptsize$\pm$0.06} & 58.73{\scriptsize$\pm$1.58} & 86.24{\scriptsize$\pm$2.87}
& 89.75{\scriptsize$\pm$0.28} & 90.69{\scriptsize$\pm$0.08} & 56.42{\scriptsize$\pm$1.95} & 85.43{\scriptsize$\pm$3.16} \\
\midrule
TGAT-TIGPrompt
& 69.21{\scriptsize$\pm$8.88} & 67.70{\scriptsize$\pm$9.64} & 73.90{\scriptsize$\pm$6.68} & 51.38{\scriptsize$\pm$2.72}
& 59.54{\scriptsize$\pm$1.41} & 78.45{\scriptsize$\pm$1.44} & 51.69{\scriptsize$\pm$1.24} & 69.71{\scriptsize$\pm$4.16}
& 49.52{\scriptsize$\pm$0.85} & 65.66{\scriptsize$\pm$2.68} & 51.58{\scriptsize$\pm$4.02} & 48.34{\scriptsize$\pm$3.28} \\
TGN-TIGPrompt
& 44.80{\scriptsize$\pm$5.45} & 63.75{\scriptsize$\pm$5.60} & 55.42{\scriptsize$\pm$3.60} & 50.84{\scriptsize$\pm$2.75}
& 82.04{\scriptsize$\pm$2.03} & 83.26{\scriptsize$\pm$2.38} & 65.00{\scriptsize$\pm$4.73} & 86.25{\scriptsize$\pm$2.43}
& 81.75{\scriptsize$\pm$1.97} & 79.51{\scriptsize$\pm$2.58} & 64.98{\scriptsize$\pm$4.61} & 86.19{\scriptsize$\pm$3.06} \\
\midrule
TGAT-DyGPrompt
& \textbf{82.09}{\scriptsize$\pm$6.43} & 73.50{\scriptsize$\pm$6.47} & \underline{77.78}{\scriptsize$\pm$5.08} & \textbf{52.03}{\scriptsize$\pm$2.24}
& 69.88{\scriptsize$\pm$0.18} & 90.76{\scriptsize$\pm$0.09} & 53.92{\scriptsize$\pm$0.97} & 72.04{\scriptsize$\pm$4.71}
& 52.58{\scriptsize$\pm$0.23} & 75.20{\scriptsize$\pm$0.17} & 53.29{\scriptsize$\pm$0.87} & 50.82{\scriptsize$\pm$3.67} \\
TGN-DyGPrompt
& 74.47{\scriptsize$\pm$3.44} & \underline{74.00}{\scriptsize$\pm$3.10} & 69.06{\scriptsize$\pm$3.89} & \underline{51.97}{\scriptsize$\pm$2.16}
& \underline{94.33}{\scriptsize$\pm$0.12} & \underline{96.82}{\scriptsize$\pm$0.06} & \underline{70.17}{\scriptsize$\pm$0.75} & \underline{87.02}{\scriptsize$\pm$1.63}
& \underline{92.22}{\scriptsize$\pm$0.19} & \underline{95.69}{\scriptsize$\pm$0.08} & \underline{69.77}{\scriptsize$\pm$0.66} & \underline{87.63}{\scriptsize$\pm$1.97} \\
\midrule
\model{}
& \underline{76.17}{\scriptsize$\pm$4.89} & \textbf{75.60}{\scriptsize$\pm$4.75} & \textbf{79.15}{\scriptsize$\pm$3.20} & 50.27{\scriptsize$\pm$0.48}
& \textbf{95.70}{\scriptsize$\pm$0.04} & \textbf{97.30}{\scriptsize$\pm$0.09} & \textbf{84.05}{\scriptsize$\pm$0.31} & \textbf{94.40}{\scriptsize$\pm$0.21}
& \textbf{93.56}{\scriptsize$\pm$0.05} & \textbf{96.33}{\scriptsize$\pm$0.09} & \textbf{83.80}{\scriptsize$\pm$0.31} & \textbf{93.93}{\scriptsize$\pm$0.18} \\
\bottomrule
\end{tabular}%
}
\end{table*}

\subsection{Ablation Study}
\begin{table}[t]
\centering
\caption{Ablation study on Reddit and MOOC evaluated by AUC-ROC (\%).}
\label{tab:ablation}
\resizebox{\linewidth}{!}{%
\begin{tabular}{lcccccc}
\toprule
Method & \multicolumn{3}{c}{Reddit} & \multicolumn{3}{c}{MOOC} \\
\cmidrule(lr){2-4}\cmidrule(lr){5-7}
 & LP-Trans & LP-Indu & NC & LP-Trans & LP-Indu & NC \\
\midrule
\textbf{\model{}} & \textbf{97.30} & \textbf{96.33} & \textbf{75.60} & \textbf{84.05} & \textbf{83.80} & \textbf{79.15} \\
\model{} (w/o Topology) & 97.22 & 96.31 & 73.06 & 78.11 & 77.77 & 75.57 \\
\model{} (w/o Expert Prompt) & 95.49 & 93.89 & 73.53 & 77.17 & 77.10 & 75.61 \\
\model{} (w/o GE) & 95.71 & 94.24 & 72.53 & 76.63 & 77.02 & 76.75 \\
\model{} (w/o Balance) & 93.42 & 91.92 & 74.68 & 72.74 & 72.63 & 77.10 \\
\model{} (w/o Expert) & 87.29 & 83.98 & 66.43 & 56.32 & 56.27 & 78.41 \\
\bottomrule
\end{tabular}%
}
\end{table}

We ablate the key components of \model{} on Reddit and MOOC (Table~\ref{tab:ablation}). Removing any module degrades performance relative to the full model, indicating that each design choice contributes.
(i) The strongest degradation occurs under w/o Expert (i.e., neither multi-expert routing nor multi-$\kappa$  specialization). LP drops sharply, approaching chance level (e.g., MOOC LP-Old 84.05→56.32 ), indicating that the curvature-specialized expert mixture is the main contributor to interaction modeling. The NC impact is dataset-dependent: Reddit NC falls substantially (75.60→66.43 , the largest among NC ablations), whereas MOOC NC remains nearly unchanged (79.15→78.41 ).
(ii) w/o Balance yields the second-largest LP degradation on both datasets, indicating that gate-load balancing is necessary to keep multi-$\kappa$  routing effective and avoid expert under-utilization.
(iii) w/o GE (i.e., disabling $\mathcal{L}_{\mathrm{ortho}}$ and $\mathcal{L}_{D}$) and w/o Expert Prompt both cause moderate, consistent drops, showing that geometric regularization and per-expert prompts are useful.
(iv) w/o Topology leaves LP nearly unchanged on Reddit ($<$ 0.1), but causes clear drops on MOOC ($\approx$ 6 points) and consistently hurts NC on both datasets, implying that topology encoding supplies important condition/routing signals. Overall, \model{} benefits from the synergy of topology-conditioned routing,
mixed-curvature experts with specialized prompts, balanced gating, and geometric regularization.

However, w/o Expert simultaneously removes expert capacity, routing, and curvature specialization; it therefore cannot isolate whether the gains come from MoE capacity or from mixed-curvature geometry.

\begin{table}[t]
\centering
\caption{Geometry attribution on Reddit and MOOC by AUC-ROC (\%).
All variants keep the same expert count and gating; only the curvature
configuration changes.}
\label{tab:ablation_geometry}
\resizebox{\linewidth}{!}{%
\begin{tabular}{lcccccc}
\toprule
Method & \multicolumn{3}{c}{Reddit} & \multicolumn{3}{c}{MOOC} \\
\cmidrule(lr){2-4}\cmidrule(lr){5-7}
 & LP-Trans & LP-Indu & NC & LP-Trans & LP-Indu & NC \\
\midrule

\textbf{\model{} (Mixed Curvature)} & \textbf{97.30} & \textbf{96.33} & \textbf{75.60} & \textbf{84.05} & \textbf{83.80} & \textbf{79.15} \\
Single Curvature & 95.64 & 94.23 & 70.65 & 75.68 & 75.74 & 78.56 \\
Euclidean MoE & 94.56 & 92.84 & 69.18 & 68.92 & 69.02 & 76.57 \\
\bottomrule
\end{tabular}%
}
\end{table}

\paragraph{Necessity of mixed curvature.}
To disentangle MoE capacity from geometric diversity, we further compare
three geometry configurations with matched expert count and routing
(Table~\ref{tab:ablation_geometry}): Euclidean MoE (all $\kappa{=}0$),
Single Curvature (shared non-Euclidean $\kappa$), and Mixed Curvature
(\model{}).
Euclidean MoE already outperforms w/o Expert on LP, confirming that the
expert mixture itself is beneficial; yet it remains clearly below
\model{} on both datasets.
Replacing all-Euclidean experts with a shared curvature improves LP/NC,
showing that non-Euclidean geometry helps beyond MoE capacity alone.
\model{} further improves over Single Curvature,
indicating that the gains are not merely from using a curved space, but
from \emph{multi-$\kappa$} specialization across experts.

\subsection{Expert Analysis}
\label{sec:expert_gate_time}
\begin{figure}[t]
  \centering
  \includegraphics[width=0.9\linewidth]{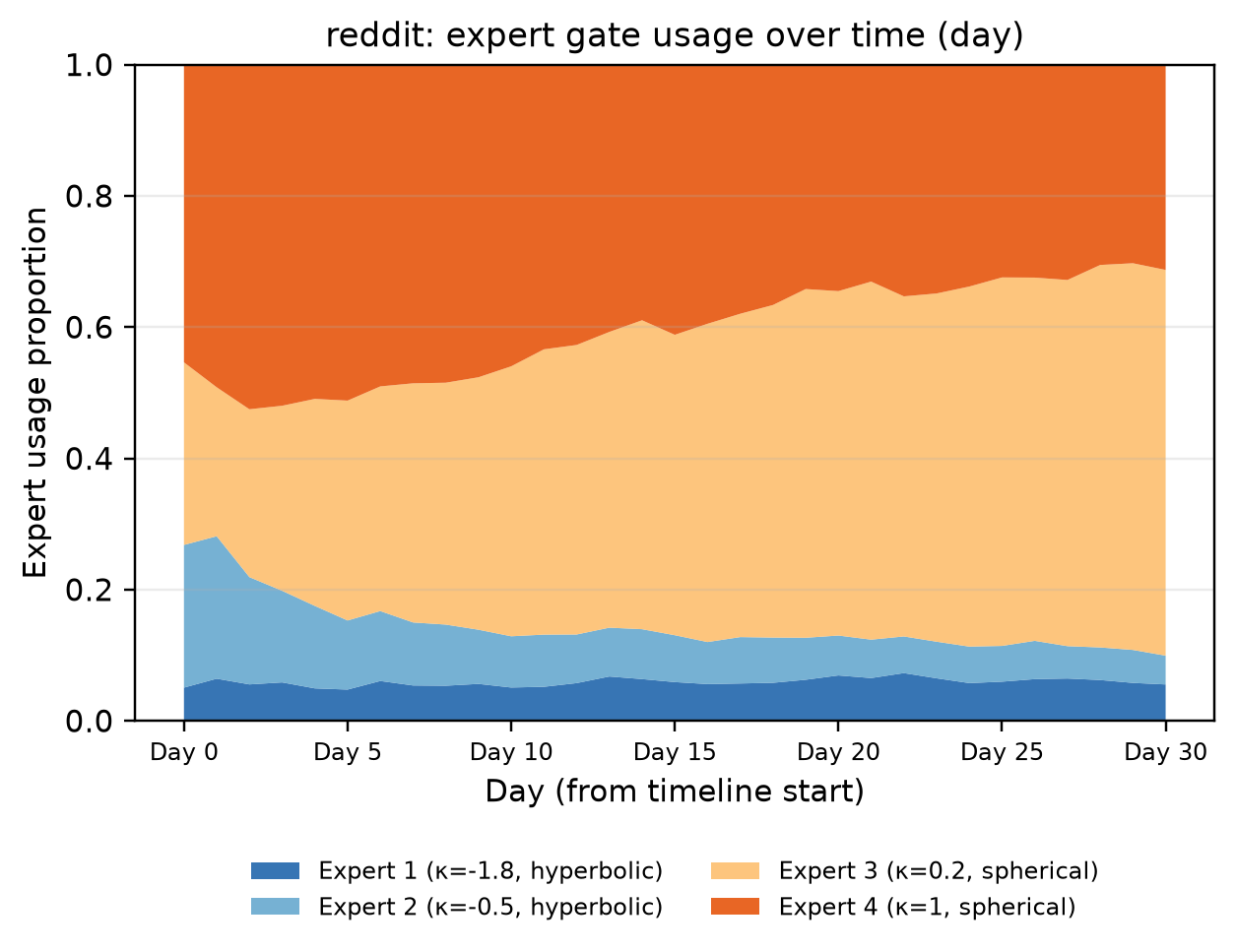}
  \caption{Reddit expert gate usage over 31 days.
    Each stacked region is the daily mean gate weight of one curvature expert.
    Routing mass shifts from $\kappa{=}1.0$ toward $\kappa{=}0.2$, while hyperbolic experts remain minority.}
  \label{fig:REDDIT_GATE}
\end{figure}

To examine whether \model{} adapts expert allocation over the interaction timeline,
we freeze a pretrained model and run causal inference on Reddit,
recording gate weights for every edge. Edges are then aggregated by day
to obtain the mean usage of each curvature expert (Fig.~\ref{fig:REDDIT_GATE}). For clearer visualization, we use a four-expert probe ($\kappa{\in}\{-1.8,-0.5,0.2,1\}$, Top-$K{=}1$), not the main Reddit setting. The resulting stacked usage reveals a clear temporal shift. Early in the timeline, the high-curvature spherical expert ($\kappa{=}1$) accounts for a substantial share of routing decisions, while hyperbolic experts ($\kappa{\in}\{-1.8,-0.5\}$)
remain secondary. Over the subsequent days, usage gradually concentrates on the mild
spherical expert ($\kappa{=}0.2$), which becomes dominant by the end of the period;
meanwhile, both $\kappa{=}1$ and the hyperbolic experts decline. This trend indicates that the learned gate does not collapse to a fixed expert mixture. Instead, it redistributes mass across complementary curvature regimes as interactions progress, supporting the necessity of multi-$\kappa$ MoE routing for temporal graphs. 


\subsection{Hyperparameter Sensitivity}

\begin{figure}[t]
  \centering
  \includegraphics[width=0.4\textwidth]{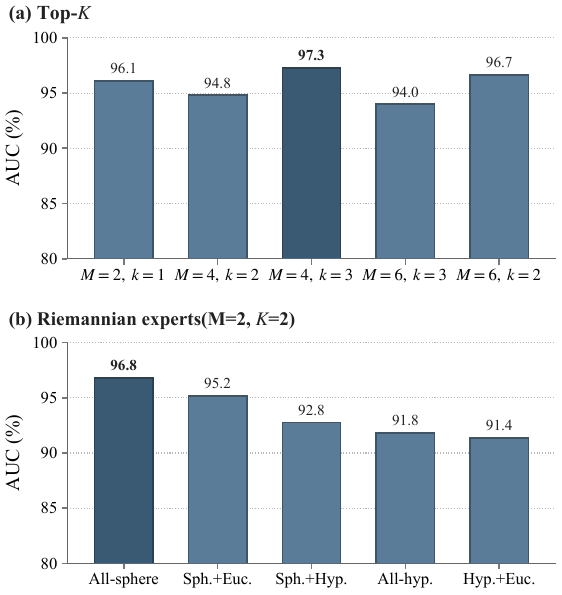}
  \caption{Sensitivity analysis on Reddit (LP).
  (a)~AUC (\%) under different Top-$K$ settings $(M,K)$, where each expert's
  $\kappa$ is drawn from a pre-defined multi-type pool; the default configuration is shown in bold.
  (b)~AUC (\%) of explicit two-expert $\kappa$ compositions under fixed
$M{=}2,K{=}2$.}
\label{fig:topk_kappa}
\end{figure}

Sensitivity to Top-$K$ and curvature.
Fig.~\ref{fig:topk_kappa} examines Top-$K$ routing and the Riemannian experts on Reddit. In Fig.~\ref{fig:topk_kappa}(a), while varying $(M,K)$, each expert's curvature $\kappa$ is
drawn from a pre-defined multi-type pool (positive, near-zero,
and negative values). AUC stays in a narrow band (94.0--97.3), showing that Top-$K$ is robust under such randomized $\kappa$ assignments. Fig.~\ref{fig:topk_kappa}(b) fixes $(M{=}2,K{=}2)$
and compares different $\kappa$ experts, where the spread is larger
(91.4--96.8): the all-spherical experts performs best, while
hyperbolic-containing experts are weaker, consistent with the learned
routing behavior on Reddit, where gate mass concentrates on the mild
spherical expert while retaining non-trivial weight on hyperbolic experts (Fig.~\ref{fig:REDDIT_GATE}). This spread far exceeds that across Top-$K$ settings in (a), indicating that the geometric composition of the expert is the dominant factor. Since the optimal expert is dataset-dependent and unknown a priori, committing to any fixed geometry is brittle. Overall, the curvature composition of the expert bank is more consequential than the exact Top-$K$ setting---and across both panels, diversity, rather than any single geometry, is the robust choice.

\section{Conclusion}
In this paper, we identify geometry under-adaptation in dynamic graph prompting and propose CurvPrompt, a topology-routed Riemannian mixture-of-experts framework. CurvPrompt utilizes a bank of curvature-diverse experts with learnable prompts and routes each node--time instance via a topology-aware gate, employing soft Top-$K$ routing during pre-training and hard Top-$K$ routing during downstream adaptation for parameter-efficient geometry adaptation. Extensive experiments demonstrate that CurvPrompt achieves the best few-shot link prediction and strong node classification across four benchmarks. 

\bibliography{references}


\appendix

\section{Appendix A Additional Experimental Details}

\subsection{A.1 Environment}

All experiments are conducted in the following environment:
\begin{itemize}
    \item Operating system: Ubuntu 22.04 LTS
    \item CPU: 13th Gen Intel Core i9-13900K
    \item GPU: 2$\times$ NVIDIA GeForce RTX 3090 (24\,GB)
    \item Memory: 128\,GB RAM
\end{itemize}

\subsection{A.2 Datasets details}

Table~\ref{tab:datasets} summarizes the datasets used in our experiments.
Wikipedia, Reddit, and MOOC are bipartite interaction networks with binary labels and 172-dimensional features over a 30-day span,
while Genre is a larger-scale user--genre network from TGB with multi-class labels and a much longer time span.

\subsection{A.3 Sensitivity to Distortion and Balance Weights}

\begin{figure}[t]
    \centering
    \includegraphics[width=0.5\textwidth]{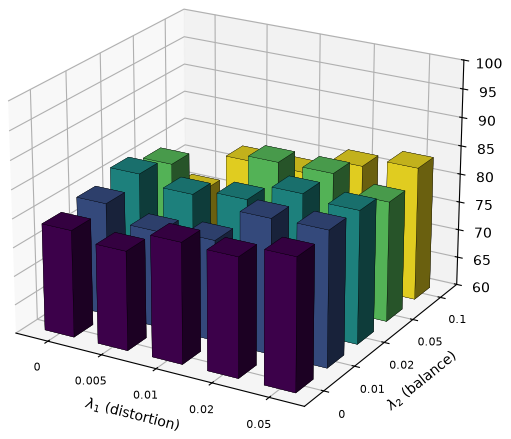}
    \caption{Sensitivity of the auxiliary loss weights on MOOC (LP-Trans AUC, \%).
    We sweep $\lambda_1$ (distortion) and $\lambda_2$ (balance) over a $5\times5$ grid while keeping all remaining hyperparameters fixed.}
    \label{fig:loss_weight_sens}
\end{figure}

Figure~\ref{fig:loss_weight_sens} illustrates the effect of varying
$(\lambda_1,\lambda_2)$
on the MOOC dataset.
Performance is sensitive to the relative scale of the two auxiliary weights.
The best performance is achieved with moderate distortion regularization
($\lambda_1\approx0.02$)
combined with a matched balance weight
($\lambda_2\in\{0.01,0.02,0.05\}$),
which is also adopted as the default configuration.
In contrast, mismatched settings—particularly a very small distortion weight paired with a larger balance weight, or a large balance weight without sufficient distortion regularization—lead to noticeable performance degradation.
These results indicate that both auxiliary objectives are beneficial when properly balanced, although the method is not insensitive to arbitrary weight choices.

\begin{table}
    
\centering
\caption{Summary of datasets.}
\label{tab:datasets}
\small
\setlength{\tabcolsep}{3.5pt}
\begin{tabular}{lrrrrrr}
\toprule
Dataset & Nodes & Edges & Classes & Labels & Feat.\ dim & Time span \\
\midrule
Wikipedia & 9,227 & 157,474 & 2 & 217 & 172 & 30 days \\
Reddit & 11,000 & 672,447 & 2 & 366 & 172 & 30 days \\
MOOC & 7,144 & 411,749 & 2 & 4,066 & 172 & 30 days \\
Genre & 1,505 & 17,858,395 & 474 & 984 & 86 & 1,500 days \\
\bottomrule
\end{tabular}
\end{table}

\end{document}